\documentclass{article}

\usepackage{microtype}
\usepackage{graphicx}
\usepackage{subfigure}
\usepackage{booktabs} % for professional tables

\usepackage{hyperref}

\usepackage[accepted]{icml2019}

\icmltitlerunning{Neural Separation of Observed and
Unobserved Distributions}

\begin{document}

\twocolumn[
\icmltitle{Neural Separation of Observed and
Unobserved Distributions}



\icmlsetsymbol{equal}{*}

\begin{icmlauthorlist}
\icmlauthor{Tavi Halperin}{huji}
\icmlauthor{Ariel Ephrat}{google}
\icmlauthor{Yedid Hoshen}{huji,fair}
\end{icmlauthorlist}

\icmlaffiliation{huji}{Department of Computer Science, The Hebrew University of Jerusalem, Jerusalem, Israel}
\icmlaffiliation{google}{Google Research}
\icmlaffiliation{fair}{Facebook AI Research}

\icmlcorrespondingauthor{Tavi Halperin}{tavi.halperin@mail.huji.ac.il }

\icmlkeywords{Machine Learning, ICML}

\vskip 0.3in
]



\printAffiliationsAndNotice{}  % leave blank if no need to mention equal contribution

\begin{abstract}

Separating mixed distributions is a long standing challenge for machine learning and signal processing. Most current methods either rely on making strong assumptions on the source distributions or rely on having training samples of each source in the mixture. In this work, we introduce a new method---Neural Egg Separation---to tackle the scenario of extracting a signal from an unobserved distribution additively mixed with a signal from an observed distribution. Our method iteratively learns to separate the known distribution from progressively finer estimates of the unknown distribution. In some settings, Neural Egg Separation is initialization sensitive, we therefore introduce Latent Mixture Masking which ensures a good initialization. Extensive experiments on audio and image separation tasks show that our method outperforms current methods that use the same level of supervision, and often achieves similar performance to full supervision.

\end{abstract}

\section{Introduction}
\label{sec:Introduction}

Humans are remarkably good at separating data coming from a mixture of distributions, e.g. hearing a person speaking in a crowded cocktail party. Artificial intelligence, on the the hand, is far less adept at separating mixed signals. This is an important ability as signals in nature are typically mixed, e.g. speakers are often mixed with other speakers or environmental sounds, objects in images are typically seen along other objects as well as the background. Understanding mixed signals is harder than understanding pure sources, making source separation an important research topic.    

Most previous work focused on the following settings:

\textit{Full supervision:} The learner has access to a training set including samples of mixed signals $\{y\}\in\cal Y$ as well as the ground truth sources of the same signals $\{b\}\in\cal B$ and $\{x\}\in\cal X$ (such that $y = x + b$). Having such strong supervision is very potent, allowing the learner to directly learn a mapping from the mixed signal $y$ to its sources $(x, b)$. Obtaining such strong supervision is nearly never possible, as it requires knowing for each input mixture $y$, its exact separate signals $(x, b)$. This information is rarely available in a single microphone setting. 

\textit{Synthetic full supervision:} The learner has access to a training set containing samples from the mixed signal $\{y\}\in\cal Y$ as well as samples from all source distributions $\{b\}\in\cal B$ and $\{x\}\in\cal X$. The learner however does not have access to paired sets of the mixed and unmixed signal ground truth (that is for any given $y$ in the training set, its separate $b$ and $x$ are unknown). This supervision setting is more realistic than the fully supervised case, and occurs when each of the source distributions can be sampled in isolation (e.g. we can record a violin and piano separately in a studio and can thus obtain unmixed samples of each of their distributions). It is typically solved by learning to separate synthetic mixtures $b + x$ of independently sampled $b$ and $x$.   

\textit{No supervision:} The learner only has access to training samples of the mixed signal $\cal Y$ but not to sources $\cal B$ and $\cal X$. Although this settings puts the least requirements on the training dataset, it is a hard problem and can be poorly specified in the absence of strong assumptions and priors. It is generally necessary to make strong assumptions on the properties of the component signals (e.g. smoothness, low rank, periodicity) in order to make progress in separation. This limits the applicability of such methods.

In this work we concentrate on the \textit{semi-supervised} setting: unmixing of signals in the case where the mixture $\cal Y$ consists of a signal coming from an unobserved distribution $\cal X$ and another signal from an observed distribution $\cal B$ (i.e. the learner has access to a training set of clean samples such that $\{b\}\in\cal B$ along with different mixed samples $\{y'\}\in\cal Y$). One possible way of obtaining such supervision is to label every element of a signal (e.g. short waveform segment) by a label, indicating if it comes only from the observed distribution $\cal B$ or if it is a mixture of both distributions $\cal B+X$. The task is to learn a parametric function able to separate the mixed signal $y \in \cal Y$ into sources $x \in \cal X$ and $b \in \cal B$ s.t. $y = b + x$. Such supervision is more generally available than full supervision, while the separation problem becomes much simpler than when fully unsupervised.

We introduce a novel method: Neural Egg Separation (NES), consisting of iterative: $i$) estimation of samples from the unobserved distribution $\cal X$ $ii$) synthesis of mixed signals from known samples of $\cal B$ and estimated samples of $\cal X$ $iii$) training of separation function for the mixed signal. Iterative refinement of the estimated samples of $\cal X$ significantly increases the accuracy of the learned separation function. The method is named Neural Egg Separation, as it is akin to the iterative technique commonly used for separating egg whites and yolks. 

As an iterative technique, NES can be initialization sensitive. We therefore introduce another method --- Latent Mixture Masking (LMM) --- to provide NES with a strong initialization. Our method trains two deep generators end-to-end using Latent Mixtures to model the observed and unobserved sources ($\cal B$ and $\cal X$). we found that a simple initialization is sufficient when $\cal X$ and $\cal B$ are uncorrelated, whereas LMM-initialization is most important when $\cal X$ and $\cal B$ are strongly correlated such as e.g. separation of music into instruments and vocals. Initialization by LMM was found to be much more effective than by adversarial methods.

Experiments are conducted across multiple domains (image, music, voice) validating the effectiveness of our method, and its superiority over current methods that use the same level of supervision. Our semi-supervised method is often competitive with the fully supervised baseline, while making few assumptions on the nature of the component signals and requiring lightweight supervision. An analysis of the assumptions made by the method is detailed in Sec.~\ref{sec:analysis}.

\section{Previous Work}
\label{sec:prev}

\textit{Source separation:} Separation of mixed signals has been extensively researched. In this work, we focus on single channel separation. Unsupervised (blind) single-channel methods include: Robust Principal Component Analysis (RPCA) \citep{huang2012singing} and Single-channel Independent Component Analysis (ICA) \citep{davies2007source}. These methods attempt to use coarse priors about the signals such as low rank, sparsity or non-gaussianity. Hidden Markov Models (HMM) can be used as a temporal prior for longer clips \citep{roweis2001one}, however here we do not assume long clips. Supervised source separation has also been extensively researched, classic techniques often used learned dictionaries for each source e.g. Non-negative Matrix Factorization (NMF) \citep{wilson2008speech}. Recently, neural network-based separation gained popularity, usually learning a regression between the mixed and unmixed signals either directly \citep{huang2014deep}, or by regressing a multiplicative mask \citep{wang2014training, yu2017permutation}. Some methods were devised to exploit the temporal nature of long audio signal by using Reccurent Neural Networks (RNNs) \citep{mimilakis2017monaural}, in this work we concentrate on separation of short audio clips and consider such line of works as orthogonal. One related direction is Generative Adversarial Source Separation \citep{stoller2017adversarial, subakan2017generative} that uses adversarial training to match the unmixed source distributions. This is needed to deal with correlated sources for which learning a regressor on synthetic mixtures is less effective. We present an Adversarial Masking (AM) method that tackles the semi-supervised rather than the fully supervised scenario and overcomes mixture collapse issues not present in the fully supervised case. We found that non-adversarial methods perform better for the initialization task.    

The most related set of works is semi-supervised audio source separation \citep{smaragdis2007supervised, barker2014semi}, which like our work attempt to separate mixtures $\cal Y$ given only samples from the distribution of one source $\cal B$. Typically NMF or Probabilistic Latent Component Analysis (PLCA) (which is a similar algorithm with a probabilistic formulation) are used. We show experimentally that our method significantly outperforms NMF. A very early related technique is Spectral Subtraction \cite{boll1979suppression}, however it can only handle very simple unknown sources.

\textit{Disentanglement:} Similarly to source separation, disentanglement also deals with separation in terms of creating a disentangled representation of a source signal, however its aim is to uncover latent factors of variation in the signal such as style and content or shape and color e.g. \cite{denton2017unsupervised, higgins2016beta}. Differently from disentanglement, our task is separating signals rather than the latent representation. 

\textit{Generative Models:} Generative models learn the distribution of a signal directly. Classical approaches include: singular-value decomposition (SVD) for general signals and NMF \citep{lee2001algorithms} for non-negative signals. Recently several deep learning approaches dominated generative modeling including: Generative Adversarial Network (GAN) \citep{goodfellow2016deep}, Variational Autoencoder (VAE) \citep{vae} and Generative Latent Optimization (GLO) \citep{bojanowski2017optimizing}. Adversarial training (for GANs) is rather tricky and often leads to mode-collapse. GLO is non-adversarial and allows for direct latent optimization for each source making it more suitable than VAE and GAN.  

%\textit{Adversarial training:} GANs \citep{goodfellow2014generative} were proposed for the task of unconditional image generation. They consist of: i) a generator for synthesizing realistic images from noise, and ii) a discriminator trying to detect if an input image is real or fake. The generator is trained to fool the discriminator, while the discriminator attempts to detect the generator images. This paradigm has been extended \citep{kim2017learning, CycleGAN2017} to domain mapping, with a generator mapping samples from domain $\cal Y$ into fake samples from target domain $\cal B$, and a discriminator attempting to classify between real and fake $\cal X$ images. Our method also uses adversarial training to ensure that the distribution of the generated and real samples of known component have the same distribution.
%GAN training is notoriously unstable. Much recent work has been done to mitigate those issues e.g. LSGAN \citep{mao2017least}, WGAN \citep{arjovsky2017wasserstein}, Improved WGAN \citep{gulrajani2017improved} and Spectral Norm \citep{miyato2018spectral}. We have found Spectral Norm and LSGAN to significantly improve the quality of our results.

\section{Neural Egg Separation (NES)}
\label{sec:method}

In this section we present our method for separating a mixture of sources of known and unknown distributions. We denote the mixture samples $y$, the corresponding samples with the observed distribution $b$ and the samples from the unobserved distribution $x$. Our objective is to learn a parametric function $T(y)$, such that $b = T(y)$.

\textbf{Full Supervision:} In the fully supervised setting, where matching pairs ($y$, $b$) are available, this task reduces to a standard supervised regression problem, in which a parametric function $T(y)$ (typically a deep neural network) is used to directly optimize:
\begin{equation}
\label{eq:full_sup}
    T = arg\min_{T'}\sum_{(y, b)}{L_1(T'(y), b)}
\end{equation}

Mixed-unmixed pairs are usually unavailable, but in some cases it is possible to obtain a training set which includes independent samples from $\cal X$ and $\cal B$ e.g. \citep{wang2014training,yu2017permutation}. Methods typically randomly sample $x \in \cal X$ and $b \in \cal B$ sources and synthetically create mixtures $y'= x + b$. The synthetic pairs $(y', b)$ can then be used to optimize Eq.~\ref{eq:full_sup}. Note that in cases where $\cal X$ and $\cal B$ are correlated (e.g. vocals and instrumental accompaniment which are temporally dependent), random synthetic mixtures of $x$ and $b$ might not be representative of $\cal Y$ and cause difficulty generalizing to real mixtures.

\textbf{Semi-Supervision:} In many scenarios, clean samples of both mixture components are not available. Consider for example a street musical performance. Although crowd noises without street performers can be easily observed, street music without crowd noises are much harder to come by. In this case, therefore, samples from the distribution of crowd noise $\cal B$ are available, whereas the samples from the distribution of the music $\cal X$ are unobserved. Samples from the distribution of the mixed signal $\cal Y$ i.e. the crowd noise mixed with the musical performance are also available.

The example above illustrates a class of problems for which the distribution of the mixture and a single source are available, but the distribution of another source is unknown. In such cases, it is not possible to optimize Eq.~\ref{eq:full_sup} directly due to the unavailability of paired ($y$, $b$).

\textbf{Neural Egg Separation:}  Fully-supervised optimization (as in Eq.~\ref{eq:full_sup}) is very effective when pairs of ($y$, $b$) are available. We present a novel algorithm, which iteratively solves the semi-supervised task as a sequence of supervised problems without any clean training examples of $\cal X$.

The core idea of our method is that although no clean samples from $\cal X$ are given, it is still possible to learn to separate mixtures of observed samples $b$ from distribution $\cal B$ combined with some estimates of the unobserved distribution samples $x^{t}$ (where $t$ denotes the iteration of NES). Synthetic mixtures are created by randomly sampling an approximate sample $x^{t}$ from the unobserved distribution and combining with training sample $b$, thereby creating pairs $(y^{t}, b)$ for supervised training:
\begin{equation}
\label{eq:sep_egg_synth}
y^{t} = b + x^{t}
\end{equation}

 Note that the empirical distribution of synthetic mixtures ${\cal Y}^t$ might differ from the real mixture sample distribution $\cal Y$. We show empirically that there are interesting cases for which it converges towards the correct distribution: ${\cal Y}^t \rightarrow \cal Y$.

During each iteration of NES, a neural separation function $T^{t+1}(y^t)$ is trained on the created pairs by optimizing the following term:

\begin{equation}
\label{eq:sep_egg_opt}
    T^{t+1} = arg\min_{T'}\sum_{(y^t, b)}{L_1(T'(y^t), b)}
\end{equation}

At the end of each iteration, the separation function $T^t()$ can be used to approximately separate the training mixture samples $y$ into their sources:

\begin{equation}
\label{eq:sep_egg_approx}   
    x^t = y - T^t(y)  % b^t = T^t(y), x^t = y - b^t
\end{equation}

The refined samples $x^t \in {\cal X}^t$ are used for creating synthetic pairs for training $T^{t+1}(y^t)$ in the next iteration (as in Eq.~\ref{eq:sep_egg_opt}).

The above method relies on having an estimate of the unobserved distribution samples as input to the first iteration (${\cal X}^0$). One simple scheme is to initialize the estimates of the unobserved distribution samples in the first iteration as $x^0 = c \cdot y$, where $c$ is a constant fraction (typically 0.5). Although this initialization is very naive, we show that it performs well where the sources are independent. More advanced initializations will be discussed below.

At test time, separation is carried out by applying the trained separation function $T()$ (exactly as in Eq.~\ref{eq:sep_egg_approx}).

\begin{algorithm}[tb]
   \caption{Neural Egg Separation (NES)}
   \label{alg:nes}
   \begin{algorithmic}
   \STATE {\bfseries Input:} samples of: mixture  $\{y\}$, observed source $\{b\}$
   \STATE {\bfseries Initialize:} synthetic unobservable samples with \\ $x^0 \leftarrow c \cdot y$ or using AM or LMM
%   \STATE {\bfseries Initialize:} $T()$ with random weights
   \WHILE{$t < N$}
   \STATE {Initialize $T()$ with random weights}
    \STATE{Synthesize mixtures $y^t = b + x^t$ for all $b$ in $\cal B$ with randomly sampled $x^t$}
   \STATE{Optimize separation function for $P$ epochs:
   $T^{t+1} = arg\min_{T'}{\sum_{(y^t, b)}  L_1(T'(y^t), b)}$}
   \STATE{Update estimates of unobserved distribution samples: $x^{t+1} = y - T^{t+1}(y)$}
  \ENDWHILE

\end{algorithmic}
\end{algorithm}

% \textcolor{blue}{I'm not happy with the algorithm placement---it's confusing. Can we force it to be inline?} \textcolor{red}{Better?}

Our full algorithm is described in Alg.~\ref{alg:nes}. For optimization, we use SGD using ADAM update with a learning rate of $0.001$. In total we perform $N=10$ iterations, each consisting of optimization of $T$ and estimation of $x^t$, $P=25$ epochs are used for each optimization of Eq.~\ref{eq:sep_egg_opt}.

\textbf{Latent Mixtures:} NES is an iterative method and relies on having a good initialization. It does not take into account correlation between $\cal X$ and $\cal B$ e.g. vocals and instrumental tracks are highly related, whereas randomly sampling pairs of vocals and instrumental tracks is likely to synthesize mixtures quite different from $\cal Y$.

We present our method Latent Mixtures (LM), which separates mixtures by a distributional constraint enforced via latent generative modeling of the source signals. The method uses some latent optimization ideas from GLO \citep{bojanowski2017optimizing}. The novelty of LM is using mixtures of GLO models for separation, which has not been done before. LM training consists of two stages. We first learn a generator $G_B()$, with which for every observed training sample $b$ from $\cal B$,  a latent code $z_b$ can be found such that $b$ is reconstructed by the generator: $b=G_B(z_b)$. We learn end-to-end both the parameters of the generator $G_B()$ as well as a latent code $z_b$ for every training sample $b$. The per-sample latent codes are found by direct gradient descent over the values of $z_b$ (similar to word embeddings), rather than by a feedforward encoder. This stage is equivalent to GLO. The optimization is given by:

\begin{equation}
\label{eq:glo}
arg\min_{z_b, G_B} \sum_{b \in \cal B}\ell(G(z_b), b)
\end{equation}

Given the learned generator $G_B(z)$ for the $\cal B$ distribution, we learn generator $G_X(z)$ for the unobserved distribution $\cal X$. The idea is that every $\cal Y$ domain training sample $y$, is described by mixing a $\cal B$ domain signal generated by $G_B(z^B_y)$ as well as a $\cal X$ domain signal generated by $G_X(z^X_y)$. Note that we do not know the actual source $b$ but only have a generative model prior for the $\cal B$ distribution. As we have already learned $G_B()$ in the previous stage, we just need to learn $G_X()$ as well as the per $y$ sample latent codes $z^X_y$ and $z^B_y$. The optimization function is therefore:
\begin{equation}
\label{eq:lm_train2}
arg\min_{z^X_y, z^B_y, G_X} \ell(G_B(z_y^B) + G_X(z_y^X), y)
\end{equation}

Similarly to \cite{bojanowski2017optimizing}, we found that forcing latent codes to lay in a unit ball provides important regularization. We use $\ell=L_1$ except for color images, where we found it advantageous to use a VGG perceptual loss (implementation taken from  \cite{hoshen2018nam}). % Note that the two optimization stages can be performed end-to-end with similar results, however careful weighting of the loss functions is necessary in the joint case. In our experiments we do not jointly optimize, but instead optimize eq.~\ref{eq:glo} and then eq.~\ref{eq:lm_train2}, and therefore never have to deal with weighting the losses.

Once $G_B()$ and $G_X()$ are trained, we infer the latent codes for a test mixture by:
\begin{equation}
\label{eq:glom_infer}
arg\min_{z^X_y, z^B_y} \ell(G_B(z_y^B) + G_X(z_y^X), y)
\end{equation}

Our estimate for the sources is then:
\begin{equation}
\label{eq:glom_est}
\tilde{b}=G_B(z_y^B)~~~~~\tilde{x}=G_X(z_y^X)
\end{equation}

\textbf{Masking Function:} In additive separation tasks, the mixed signal $y$ is the sum of two positive signals $x$ and $b$. Instead of synthesizing the new sample, we can learn a neural network separation mask $m(y)$ that specifies the fraction of the signal which comes from $\cal B$ at each pixel. The attractive feature of the mask is always being in the range $[0,1]$ (in the case of positive additive mixtures of signals). Even a constant mask will preserve all signal gradients (at the cost of introducing spurious gradients). Mathematically this can be written as:

\begin{equation}
\label{eq:semi_frac_b}
    T(y) = y \odot m(y)
\end{equation}

For NES (and baseline AM described below), we implement the mapping function $T(y)$ using the element-wise product of the masking function and the mixture signal: $y \odot m(y)$. In practice, we find that learning a masking function yields much better results than synthesizing the signal directly (in line with other works e.g. \cite{wang2014training,gabbay2017seeing}).

LM does not provide a way for learning the mask directly. We refine its estimate by computing an effective mask from the element-wise ratio of estimated sources. We name the combination of LM and the post-processing masking operation, Latent Mixture Masking (LMM):
\begin{equation}
\label{eq:glom_mask}
m_{LMM}(y)=\frac{G_B(z^B_y)}{G_B(z^B_y)+G_X(z^X_y)}
\end{equation}

\textbf{Initializing Neural Egg Separation by LMM:} We devise the following method: $i$) Train LMM on the training set and infer the mask for each mixture. This is operated on images or mel-scale spectrograms at $64 \times 64$ resolution $ii$) For audio: upsample the mask to the resolution of the high-resolution linear spectrogram and compute an estimate of the $\cal X$ source linear spectrogram on the training set $iii$) Run NES on the observed $\cal B$ and estimated $\cal X$. We find experimentally that this initialization scheme improves NES to the point of being competitive with fully-supervised training.

\section{Experiments}
\label{sec:exp}

To evaluate the performance of our method, we conducted experiments on distributions taken from multiple real-world domains: images, speech and music, in cases where the two signals are correlated and uncorrelated.

\begin{figure*}
\label{fig:sbim}
  \centering
  
	\caption{A Qualitative Separation Comparison on Mixed Bag and Shoe Images}
    \begin{tabular}{cccccccc}
    \toprule
    Const & NMF & AM & LMM &  NES & LMM+NES & Sup & GT\\
%    \midrule
%    \includegraphics[width=0.08810\linewidth]{ims0/im_0.png} &
%    \includegraphics[width=0.08810\linewidth]{ims0/im_1.png} &
%    \includegraphics[width=0.08810\linewidth]{ims0/im_2.png} &
%    \includegraphics[width=0.08810\linewidth]{ims0/im_3.png} &
%    \includegraphics[width=0.08810\linewidth]{ims0/im_4.png} &
%    \includegraphics[width=0.08810\linewidth]{ims0/im_5.png} &
%    \includegraphics[width=0.08810\linewidth]{ims0/im_6.png} &
%    \includegraphics[width=0.08810\linewidth]{ims0/im_7.png} \\
    \midrule
    \includegraphics[width=0.08810\linewidth]{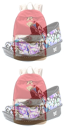} &
    \includegraphics[width=0.08810\linewidth]{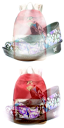} &
    \includegraphics[width=0.08810\linewidth]{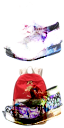} &
    \includegraphics[width=0.08810\linewidth]{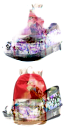} &
    \includegraphics[width=0.08810\linewidth]{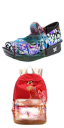} &
    \includegraphics[width=0.08810\linewidth]{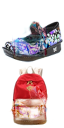} &
    \includegraphics[width=0.08810\linewidth]{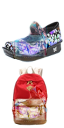} &
    \includegraphics[width=0.08810\linewidth]{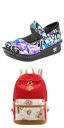} \\
  %  \midrule
  %  \includegraphics[width=0.08810\linewidth]{ims2/im_0.png} &
  %  \includegraphics[width=0.08810\linewidth]{ims2/im_1.png} &
  %  \includegraphics[width=0.08810\linewidth]{ims2/im_2.png} &
  %  \includegraphics[width=0.08810\linewidth]{ims2/im_3.png} &
  %  \includegraphics[width=0.08810\linewidth]{ims2/im_4.png} &
  %  \includegraphics[width=0.08810\linewidth]{ims2/im_5.png} &
  %  \includegraphics[width=0.08810\linewidth]{ims2/im_6.png} &
  %  \includegraphics[width=0.08810\linewidth]{ims2/im_7.png} \\

	 \bottomrule
    \end{tabular}
\end{figure*}

We evaluated our method against $3$ baseline methods: 

\textit{Constant Mask (Const):} This baseline uses the original mixture as the estimate.

\textit{Semi-supervised Non-negative Matrix Factorization (SS-NMF):} This baseline method, proposed by \cite{smaragdis2007supervised}, first trains a set of $l$ bases on the observed distribution samples $B$ by Sparse NMF \citep{hoyer2004non, kim2007sparse}. It factorizes $B = H_b W_b$, with activations $H_b$ and bases $W_b$, all matrices are non-negative. The optimization is solved using the Non-negative Least Squares solver by \cite{kim2011fast}. It then proceeds to train another factorization on the mixture $Y$ training samples with $2l$ bases, where the first $l$ bases ($W_b$) are fixed to those computed in the previous stage: $Y = H^b_y*W_b + H^x_y*W_x$. The separated sources are then: $\tilde{x} = H^x_y*W_x$ and $\tilde{b} = H^b_y*W_b$.

% \textit{Denoising  Auto-Encoder (DAE):} We improve the linear NMF baseline by introducing a deeper baseline. It first trains a deep denoising auto-encoder on all the $B$ training samples, we dub the resulting function $AE()$. It then learns a masking function $M^{AE}()$ on the mixture training samples, such that the masked $B$ signal is well reconstructed by the DAE training in the previous stage. We additionally add a prior preventing the masking function collapsing to $0$. The loss function is therefore: $L = \|AE(y_i \cdot M^{AE}(y_i)), y_i \cdot M^{AE}(y_i)\| + \lambda \|M^{AE}(y_i), p(y_i)\|$ 

\textit{Adversarial Masking (AM):} As an additional contribution, we introduce a new adversarial semi-supervised method, to improve over the shallow NMF baseline. AM trains a masking function $m()$ so that after masking, the training mixtures are indistinguishable from the distribution of source $\cal B$ under an adversarial discriminator $D()$. The loss functions (using LS-GAN \cite{mao2017least}) are given by:
\begin{equation}
\label{eq:semi_disc_d}
    D = arg\min_{D'} \sum_{y \in \cal Y}{D'(y \odot m(y))^2} + \sum_{b \in \cal B}(D'(b) - 1)^2
\end{equation}

\begin{equation}
\label{eq:semi_disc_g}
    m = arg\min_{m'} \sum_{y \in \cal Y}(D(y \odot m'(y)) - 1)^2
\end{equation}

Differently from CycleGAN \citep{CycleGAN2017} and DiscoGAN \citep{kim2017learning}, AM is not bidirectional and cannot use cycle constraints. We have found that adding a magnitude prior $L_1(m(y), 1)$ improves performance and helps prevent collapse. To partially alleviate mode collapse, we use Spectral Norm \citep{miyato2018spectral} on the discriminator. 

We evaluated our proposed methods:

\textit{Latent Mixture Masking (LMM):} LMM on mel-spectrograms or images at $64 \times 64$ resolution.

\textit{Neural Egg Separation (NES):} The NES method detailed in Sec.~\ref{sec:method}. Initializing $\cal X$ estimates using a constant ($0.5$) mask over $\cal Y$ training samples.

\textit{Initialization by Another Method (AM+NES and LMM+NES):} Initializing NES with the $\cal X$ estimates obtained by Adversarial Masking or by Latent Mixture Masking.

To upper bound the performance of our method, we also compute a \textit{fully supervised} baseline, for which paired data $(y=x+b, b)$ of $b \in \cal B$, $x \in \cal X$ and $y \in \cal Y$ are available. We train a masking function with the same architecture as used by all other regression methods to directly regress synthetic mixtures to unmixed sources. This method uses more supervision than our method and is an upper bound.

Please see the appendix for elaborate implementation details.

\subsection{Separating Mixed Images}
\label{subsec:method_image}

% In this section we evaluate the effectiveness of our method on image mixtures. We conduct experiments both on the simpler MNIST dataset and more complex Shoes and Handbags datasets.

% \subsubsection{MNIST} 
\paragraph*{MNIST}
\label{sec:exp_mnist}

We evaluate our method on image separation using the following experimental protocol. We split the MNIST dataset \citep{mnist} into two classes, the first consisting of the digits $0$-$4$ and the second consisting of the digits $5$-$9$. We conduct experiments where one source has an observed distribution $\cal B$ while the other source has an unobserved distribution $\cal X$. We use $12k$ $\cal B$ training images as the $\cal B$ training set, while for each of the other $12k$ $\cal B$ training images, we randomly sample a $\cal X$ image and additively combine the images to create the $\cal Y$ training set. We evaluate the performance of our method on $5000$ $\cal Y$ images similarly created from the test set of $\cal X$ and $\cal B$. The experiment was repeated for both directions i.e.  $0$-$4$ being $\cal B$ while $5$-$9$ in $\cal X$, as well as $0$-$4$ being $\cal X$ while $5$-$9$ in $\cal B$. 

In Tab.~\ref{tab:image}, we report our results on this task. For each experiment, the top row presents the results (peak signal-to-noise Ratio (PSNR) and structural similarity (SSIM)) on the $\cal X$ test set. Due to the simplicity of the dataset, NMF achieved reasonable performance on this dataset. LMM achieves better SSIM but worse PSNR than NMF while AM performed 1-2dB better. NES achieves much stronger performance than all other methods, achieving about 1dB worse than the fully supervised performance. Initializing NES with the masks obtained by LMM, results in similar performance to the fully-supervised upper bound. Initialization by AM achieved similar but slightly inferior performance to initialization by LMM and were omitted from the table for clarity.

\begin{table*}[t]
  \centering
      
  \caption{Image Separation Accuracy (PSNR dB/SSIM)}
  \label{tab:image}

    \begin{tabular}{llccccccc}
    \toprule
    %\multicolumn{2}{c}{Dataset}  &  \multicolumn{2}{c}{Baselines} &   \multicolumn{4}{c}{Ours} & \multicolumn{1}{c}{Supervised}   \\ 
    \cmidrule(l){1-2} \cmidrule(l){3-4} \cmidrule(l){5-8} \cmidrule(l){9-9}
	$\cal X$ & $\cal B$ & \textit{Const} & \textit{NMF} & \textit{AM} & \textit{LMM} & \textit{NES} & \textit{LMM+NES} & \textit{Supervised}   \\  
     \midrule
     $0$-$4$ & $5$-$9$ & 10.6/0.65 & 16.5/0.71 & 17.8/0.83 & 15.1/0.76 & 23.4/\textbf{0.95} & \textbf{23.9/0.95} & 24.1/0.96 \\
     $5$-$9$ & $0$-$4$ & 10.8/0.65 & 15.5/0.66 & 18.2/0.84 & 15.3/0.79 & 23.4/\textbf{0.95} & \textbf{23.8/0.95} & 24.4/0.96 \\
     Bags & Shoes & 6.9/0.48 & 13.9/0.48 & 15.5/0.67 & 15.1/0.66 & 22.3/0.85 & \textbf{22.7/0.86} & 22.9/0.86 \\
     Shoes & Bags & 10.8/0.65 & 11.8/0.51 & 16.2/0.65 & 14.8/0.65  & 22.4/0.85 & \textbf{22.8/0.86} & 22.8/0.86 \\
	 \bottomrule
    \end{tabular}
\end{table*}

% \begin{table*}[t]
%   \centering
      
%   \caption{Speech Separation Accuracy (PSNR dB)}
%   \label{tab:speech}

%     \begin{tabular}{llccccccc}
%     \toprule

%     \cmidrule(l){1-2} \cmidrule(l){3-4} \cmidrule(l){5-8} \cmidrule(l){9-9}
% 	$\cal X$ & $\cal B$ & \textit{Const} & \textit{NMF} & \textit{AM} & \textit{LMM} & \textit{NES} & \textit{FT} &  \textit{Supervised}   \\   
%     \midrule
%      Speech & Noise & 0.0 & 2.4 &  5.7 & 3.3 & \textbf{7.5} & \textbf{7.5} & 8.3 \\

% 	 \bottomrule
%     \end{tabular}

% \end{table*}

\begin{figure*}[h]
  \centering
	\caption{A qualitative comparison of mixtures of speech and noise (top and middle rows, respectively) separated by LM and LMM, as well as NES after $k$ iterations. NES($k$) denotes NES after $k$ iterations. Note that LM and LMM share the same mask (bottom row), since LMM is generated by the mask computed from LM. \label{fig:audio_viz_}}
    \begin{tabular}{ccccccc}
    \toprule
    Mix & LM & LMM & NES(1)  & NES(3)  & NES(5) & GT\\
    \midrule

\includegraphics[width=0.080\linewidth]{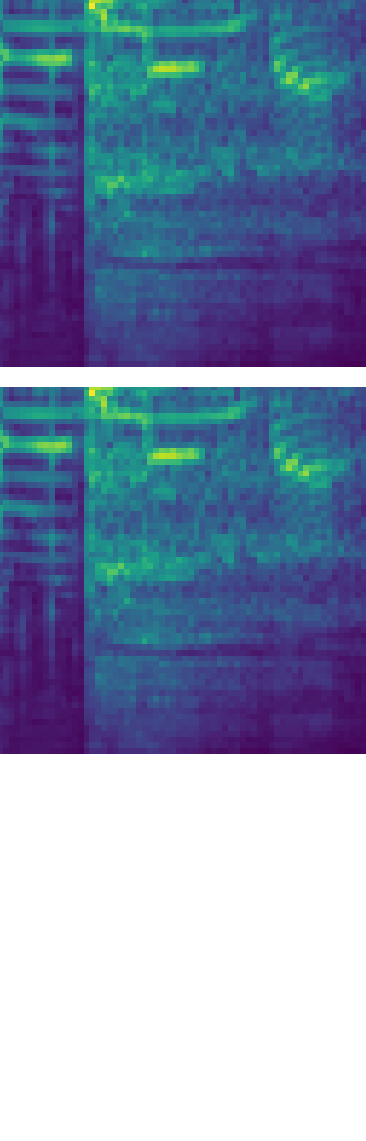} &

\includegraphics[width=0.080\linewidth]{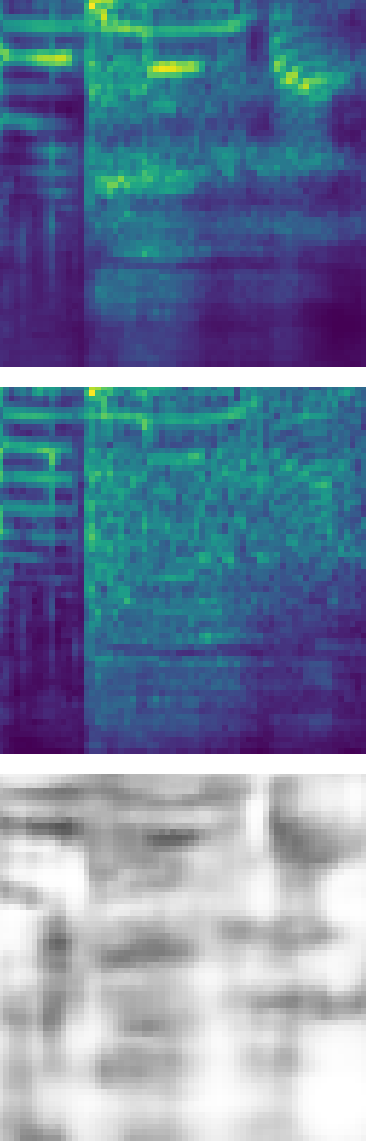} &
    \includegraphics[width=0.080\linewidth]{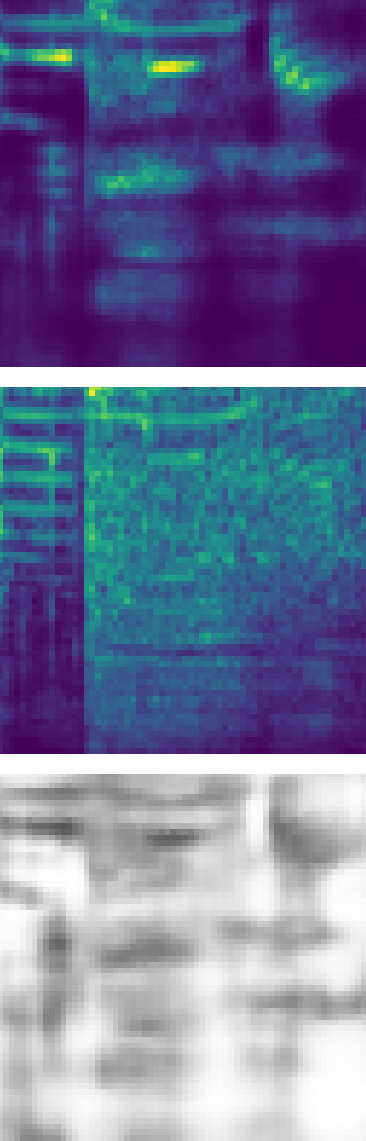} &
    \includegraphics[width=0.080\linewidth]{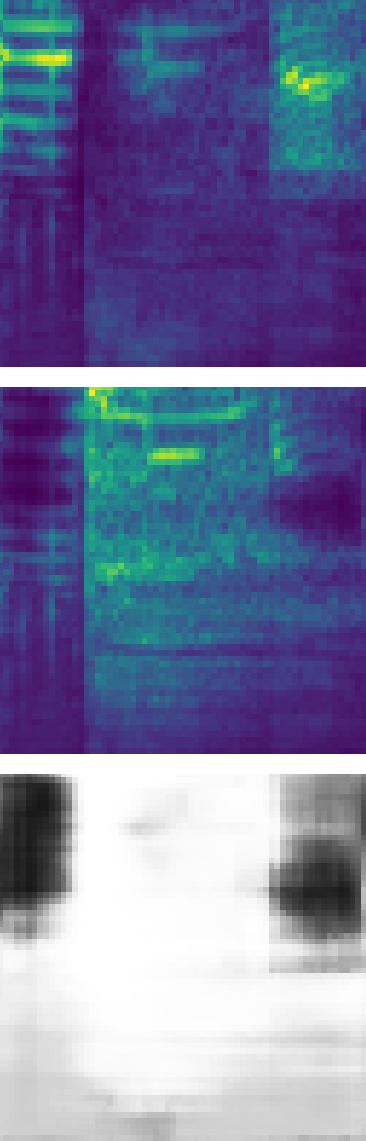} &
    \includegraphics[width=0.080\linewidth]{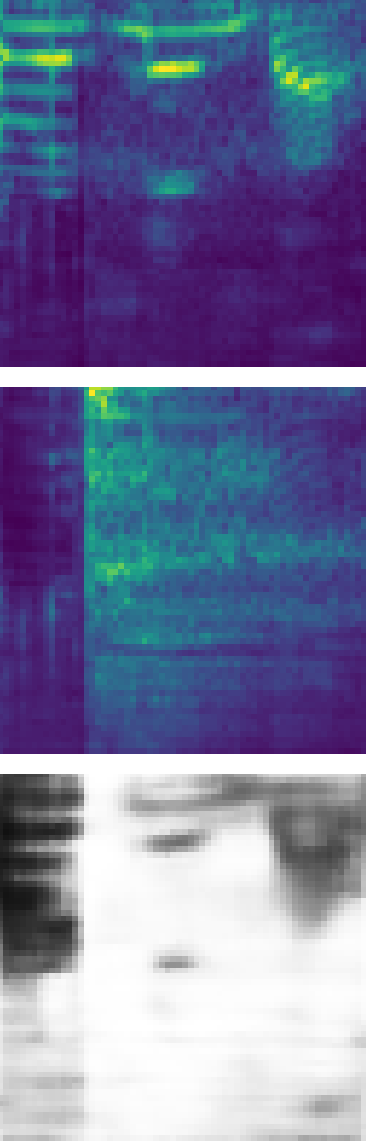} &
    \includegraphics[width=0.080\linewidth]{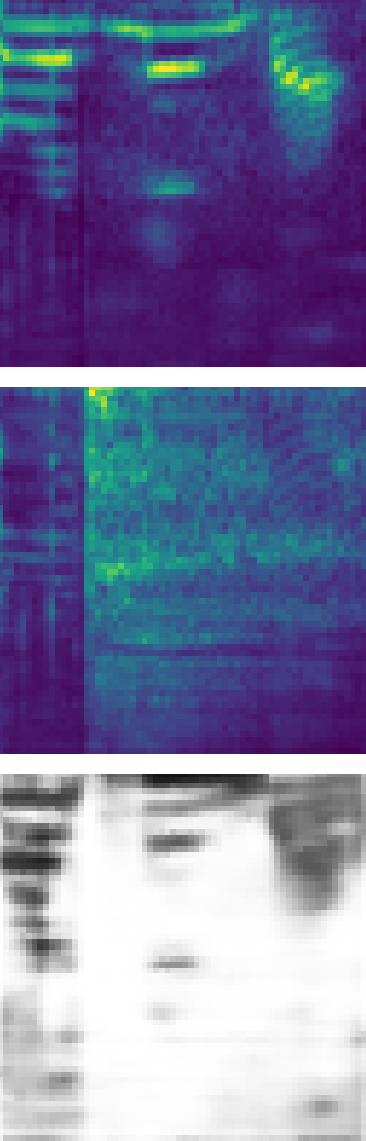} &
    \includegraphics[width=0.080\linewidth]{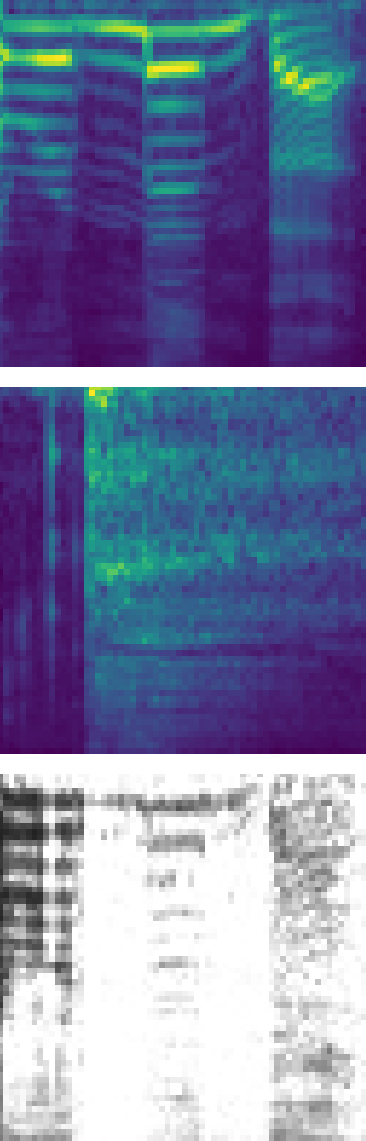} \\
    \midrule

   \\

	 \bottomrule
    \end{tabular}
\end{figure*}

\begin{table*}[h]
  \centering
      
  \caption{Audio Separation Accuracy (Median SDR dB)}
  \label{tab:audio}

    \begin{tabular}{llcccccccc}
    \toprule
    \cmidrule(l){1-2} \cmidrule(l){3-4} \cmidrule(l){5-9} \cmidrule(l){10-10}
	$\cal X$ & $\cal B$ & \textit{Const} & \textit{NMF} & \textit{AM} & \textit{LMM} & \textit{NES} &       \textit{AM+NES}          & \textit{LMM+NES}                & \textit{Supervised}     \\   
    \midrule      %         const   nmf     am   lmm   nes             am+nes         lmm+nes        supervised
     Vocals & Instrumental & 0.0  & 0.0  & 0.0 & 0.6 & 0.3          & 1.2          & \textbf{2.0} & 2.8 \\
     Drums & Instrumental  & 0.1  & -0.6 & 1.2 & 0.8 & 1.3          & 2.9          & \textbf{3.4} & 3.7 \\
     Speech & Noise        & 3.0  & 2.7  & 6.0 & 2.3 & \textbf{7.8} & 7.2          & 7.6          & 8.0 \\
     Noise & Speech        & 3.0  & 2.8  & 5.2 & 2.7 & 6.3          & \textbf{6.4} & 6.1          & 8.1 \\
	 \bottomrule
    \end{tabular}

\end{table*}

% \begin{table*}[h]
%   \centering
      
%   \caption{Music Separation Accuracy (Median SDR dB)}
%   \label{tab:music}

%     \begin{tabular}{llccccccc}
%     \toprule
%     \cmidrule(l){1-2} \cmidrule(l){3-4} \cmidrule(l){5-8} \cmidrule(l){9-9}
% 	$\cal X$ & $\cal B$ & \textit{Const} & \textit{NMF} & \textit{AM} & \textit{LMM} & \textit{NES} & \textit{FT} & \textit{Supervised}    \\   
%     \midrule
%      Vocals & Instrumental & -3.5 & 0.0 & 0.3 & 0.6 & 0.3 & \textbf{2.1} & 2.7 \\
%      Drums & Instrumental & -3.3 & -0.5 & 1.5 & 0.8 & 1.3 & \textbf{3.5} & 3.6 \\
% 	 \bottomrule
%     \end{tabular}

% \end{table*}

% \subsubsection{Bags and Shoes} 
\paragraph*{Bags and Shoes}
\label{sec:exp_bags}

To evaluate our method on more realistic images, we evaluate on separating mixtures consisting of pairs of images sampled from the Handbags \citep{edges2bags} and Shoes \citep{edges2shoes} datasets, which are commonly used for evaluation of conditional image generation methods. To create each $\cal Y$ mixture image, we randomly sample a shoe image from the Shoes dataset and a handbag image from the Handbags dataset and sum them. For the observed distribution, we sample another $5000$ different images from a single dataset. We evaluate our method both for cases when the $\cal X$ class is Shoes and when it is Handbags.

From the results in Tab.~\ref{tab:image}, we can observe that NMF failed to preserve fine details, penalizing its performance metrics. LMM (which used a VGG perceptual loss) performed much better, due to greater expressiveness. AM performance was similar to LMM on this task, as the perceptual loss and stability of training of non-adversarial models helped LMM greatly. NES performed much better than all other methods, even when initialized from a constant mask. Initialization by LMM, helped NES achieve stronger performance, nearly identical to the fully-supervised upper bound. It performed better than initialization by AM (not shown in table) which achieved $22.5/0.85$ and $22.7/0.86$ . Similar conclusions can be drawn from the qualitative comparison in the figure above.

We tested our method on a standard denoising task, where the observed source is clean images and noise unobserved. We use positively clamped Gaussian noise $\sigma=0.1$. For methods LMM/ours/supervised, we obtained PSNR: 24.4/28.5/28.5 SSIM: 0.76/0.88/0.88 on the Bags dataset and PSNR: 25.2/29.4/29.4 SSIM: 0.83/0.9/0.9 on the Shoes dataset. Our method seems to be well suited for denoising. 

\subsection{Separating Speech and Environmental Noise}
\label{subsec:method_speech}

Separating environmental noise from speech is a long standing problem in signal processing. Although supervision for both human speech and natural noises can generally be obtained, we use this task as a benchmark to evaluate our method's performance on audio signals where $\cal X$ and $\cal B$ are not dependent. This benchmark is a proxy for tasks for which a clean training set of sounds from $\cal X$ cannot be obtained e.g. for animal sounds in the wild, where
background sounds without animal noises can easily be recorded, but clean sounds made by the animal with no background sounds are unlikely to be available.

For our experiments, we use the Oxford-BBC Lip Reading in the Wild (LRW) Dataset \citep{lrw} for speech. For noise we use audio segments from ESC-$50$ \citep{esc-50}, a dataset of environmental audio recordings organized into $50$ semantic classes. Detailed description of our implementation for audio pre-processing and mask training can be found in the appendix. Separation quality is measured using Signal-to-Distortion ratio (SDR), measured using the BSS Eval toolbox \cite{BSSeval, stoter2018}.

From the speech results in Tab.~\ref{tab:audio}, we can observe that LMM performed similarly to Semi-Supervised NMF, and AM training performed about 3dB better than LMM. Due to the independence between the sources in this task, NES performed well, even when trained from a constant mask initialization. Performance was very close to the fully supervised result (when speech is unobserved). In this setting, initializing NES with the speech estimates obtained by LMM (or AM) did not yield improved performance.      

We present in Fig.~\ref{fig:audio_viz_} the results of the different methods on a mixture from the speech dataset. It can be observed that LM captures the general features of the sources, but is not able to exactly capture fine detail. The masking operation in LMM helps it recover more fine-grained details, and results in much cleaner separations. We observe that NES converged quite quickly, and results improve further with increasing iterations. Quantitative SDR results are in line with this finding. A graph of SDR for different $NES(k)$ is presented in the appendix.

\subsection{Music Separation}
\label{subsec:method_music}

Separating music into singing voice and instrumental music as well as drums separation from instrumental music has been a standard task for the signal processing community. Here our objective is to understand the behavior of our method in settings where $\cal X$ and $\cal B$ are dependent.

We use the MUSDB18 Dataset \citep{musdb18}, consisting for each music track of separate signal streams for the mixture, drums, bass, the rest of the accompaniment, and the vocals. We convert the audio tracks to mono, resample to $20480$ Hz, and then follow the same procedure as for speech to obtain input audio features.

From the music results in Tab.~\ref{tab:audio}, we can observe that NMF was the worst performer in this setting (as its simple bases do not generalize well between songs). LMM was able to do much better than NMF and was even competitive with NES on vocal-instrumental separation. Due to the dependence between the two sources and low SNR, initialization proved important for NES. Constant initialization NES performed similarly to AM and LMM. Initialization NES by LMM masks performed much better than all other methods and was competitive with the supervised baseline. LMM initialization was better than AM initialization. 

\section{Understanding the Limits of NES}
\label{sec:analysis}

This section will investigate a few scenarios under which NES is expected to converge.

\textbf{Optimal Masking:} At each iteration, we solve the following optimization problem: ${\cal{L}}^{t+1}_{NES} = L_1(m^{t+1}(y^t) \odot {y^t}, b)$, where $t$ is the iteration number, $y^t$ is a synthetic mixture consisting of the sum of a random observed sample $b \in \cal B$ and an estimated sample $x^t \in {\cal X}^t$ ($y^t = b + x^t$). At iteration $t$, the optimal mask $m^{t+1}()$ is:
\begin{equation}
\label{eq:emp_mask}
    m^{t+1}(y^t) = \frac{b}{y^t}
\end{equation}

There are several requirements for this optimization to work: $i$) Similarly to other learning-based source separation models, we assume that every mixture $y$ has a unique decomposition into separate sources $x \in \cal X$ and $b \in \cal B$. This means that the sources need to have distinct forms. This assumption is not exactly satisfied in practice, but there are cases where it is a good approximation. $ii$) We assume that the optimization method is able to find the optimal solution---despite the non-convexity of the network. The network needs to be sufficiently large to fit the data. These requirements are shared by most other deep learning works.

\textbf{Generalization from ${\cal Y}^t$ to $\cal Y$:} The objective of source separation (in our formulation) is to learn the mask yielding $m(y) = \frac{b}{y}$ for every $y \in \cal Y$. NES, however, is only able to operate on the approximate distribution ${\cal Y}^t$, where $t$ is the iteration number. To achieve the supervised performance, NES attempts to progressively improve its estimation: ${\cal Y}^t \rightarrow \cal Y$. At each iteration, the approximation of $y$ is updated using the most recent mask: $y^{t+1} = b + (y - m^{t+1}(y)y)$ (This approximation is not actually known to us as the particular $b$ component of $y$ is unknown). Convergence of the distribution can be measured by the difference between the estimated $y^t$ and the correct sample $y$. The absolute error $|e^{t+1}(y)|$ is defined below (in this discussion all operations are element-wise): 
\vspace{-0.03em}

\begin{equation}
   |e^{t+1}| = |y^{t+1} - y| = |b - m^{t+1}(y) \odot y|
\end{equation}

\vspace{-0.03em}
As $m^{t+1}$ was trained on ${\cal Y}^t$, rather than $\cal Y$, we do not know apriori how it generalizes on the true $\cal Y$ distribution. Let us consider several scenarios:

\textit{Perfect Generalization:} If $m^{t+1}$ trained on ${\cal Y}^t$ generalizes perfectly to all $y \in \cal Y$, then $m^{t+1}(y) = \frac{b}{y}$. In this case, $|e^{t+1}(y)| = |b - \frac{b}{y} \odot y| = 0$. A single iteration is therefore sufficient for convergence.

\textit{Locally Invariant $m^{t+1}$:} Instead of assuming perfect generalization, we consider the case that $m^{t+1}$ is locally invariant around $y^t$ values. The assumption is that $m^{t+1}(y) \simeq m^{t+1}(y^t) = \frac{b}{y^t}$. The error will become:
\vspace{-0.15em}

\begin{equation}
   |e^{t+1}| = |b - m^{t+1}(y^t) \odot y| = |b - \frac{b}{y^t} \odot y| = |\frac{b}{y^t} \odot (y - y^t)|
\end{equation}

\vspace{-0.05em}
We finally obtain:

\begin{equation}
\label{eq:er_invar}
   |e^{t+1}| = |\frac{b}{y^t}||e^t(y)|
\end{equation}

In this case, for $b < y^t$ (or $x^t > 0$), we obtain $|e^{t+1}| \leq |e^t(y)|$. Under these conditions the error will decrease for non-zero estimates of the unobserved signal.

\textit{Slowly Varying $m^{t+1}$:} In the general case where NES is not locally invariant around $y_t$, let us assume $m^{t+1}$ changes slowly enough so that there exists a constant $\lambda$ satisfying:

\begin{equation}
   |b - m^{t+1}(y) \odot y| \leq \lambda \dot |b - m^{t+1}(y^t) \odot y|
\end{equation}

It is possible to view $\lambda$ as a measure of generalization. $m^{t+1}$ satisfying better generalization properties will have a lower value of $\lambda$. Perfect generalization is recovered with $\lambda=0$, and invariant $m^{t+1}$ is achieved with $\lambda=1$. For general $\lambda$, the error is at most larger than Eq.~\ref{eq:er_invar} by a factor of $\lambda$:

\begin{equation}
   |e^{t+1}| \leq \lambda |\frac{b}{y^t}||e^t| 
\end{equation}

We can immediately see that convergence will occur for elements satisfying: $|\frac{b}{y^t}| < \frac{1} \lambda$. For increasing values of $\lambda$ i.e. decreasing generalization, only larger estimated values of $x^t$ (relative to $b$) will achieve decreased error.

A good initialization of $m^0$ improves its generalization ability, decreasing its value of $\lambda$. The lower $\lambda$ values increase the radius of convergence. This may explain the improved performance of better initializations.  

\section{Discussion}
\label{sec:discussion}

\textit{LMM vs. Adversarial Masking:} LMM as a stand alone technique usually performed worse than Adversarial Masking, but served as a better initialization. We speculate that mode collapse, inherent in adversarial training, makes the adversarial mask a lower bound on the $\cal X$ source distribution. LMM can result in models that are too loose (i.e. also encode samples outside of $\cal X$). But as an initialization for NES, it is better to have a model that is too loose than a model which is too tight.  

\textit{Automatic Label Extraction:} To improve sample efficiency, we hypothesize that it would be possible to label only a limited set of examples as containing the target sound and not, and to use this seed dataset to train a deep sound classifier to extract more examples from an unlabeled dataset. We leave this investigation to future work.

%\textit{More than $2$ Sources:} In the case of $L-1$ observed sources, and one unobserved, NES would work identically. In the case where only a single source was observed, NES would separate between the observed source and the mixture of unobserved sources. Separating between the mixture of unobserved sources is out-of-scope and can be attempted using blind separation methods. 

\textit{Signal-Specific Losses:} To showcase the generality of our method, we chose not to encode task specific constraints. In practical applications of our method however we believe that using signal-specific constraints can increase performance. Examples of such constraints include: repetitiveness of music \citep{rafii2011simple}, sparsity of singing voice, smoothness of natural images.

\textit{Additive and Convolution Mixtures:} In line with most of the literature, our approach separates additive mixtures. In some settings, the mixtures are convolutional. We leave the expansion of NES to this setting for future work.

\textit{Non-Adversarial Alternatives:} The good performance of LMM vs. AM on the vocals separation task, suggests that non-adversarial generative methods may be superior to adversarial methods for separation. This has also been observed in other mapping tasks e.g. the improved performance of NAM \citep{hoshen2018nam} over DCGAN \citep{dcgan}.

\section{Conclusions}
\label{sec:conc}

In this paper we proposed a novel method, Neural Egg Separation, for separating mixtures of observed and unobserved distributions. We showed that careful initialization using LMM improves results in challenging cases. Our method achieved much better performance than other methods and was usually competitive with full-supervision. Analytical results were presented to motivate the success of our method.

\section*{Acknowledgements}

We thank Lior Wolf for fruitful discussions and for coining the name ''Egg Separation''.  

\bibliography{example_paper}
\bibliographystyle{icml2019}

\end{document}